\documentclass{svproc}
\usepackage{url}

\usepackage[T1]{fontenc}
\usepackage{graphicx}
\usepackage{amsmath,amsfonts,amssymb}
\usepackage{booktabs}
\usepackage{hyperref}
\usepackage{color}
\usepackage{caption}
\usepackage{subcaption}
\usepackage{siunitx}
\usepackage{multirow}
\usepackage{float}
\usepackage{enumitem}
\usepackage{mathtools}
\usepackage{bm}
\usepackage{microtype}
\usepackage{nicefrac}
\usepackage{comment}
\usepackage{placeins}      
\usepackage{needspace}     

\graphicspath{{figures/}}
\DeclareGraphicsExtensions{.pdf,.png,.jpg}

\allowdisplaybreaks

\begin{document}
\mainmatter              
\title{Seasonal and Quantum-Inspired Models for Neutron-Monitor Time-Series Forecasting}
\titlerunning{Seasonal \& Quantum-Inspired Models for LMKS}
%
\author{Krishna Bhatia\inst{1} \and Shalini Devendrababu\inst{1} \and Srinjoy Ganguly\inst{2}}

\institute{
Quantum AI Lab, Fractal AI Research, India, \\
\email{krishna.bhatia@fractal.ai}
\and
University College London, Gower Street, London, UK
}
\authorrunning{K. Bhatia et al} 
%
%

\maketitle              

\begin{abstract}
We present a focused, reproducible study of multi-horizon forecasting on the Lomnický Štít neutron monitor (LMKS) time series. The evaluation suite spans simple seasonal baselines, modern deep sequence models and functional/quantum-inspired architectures: Seasonal Naive, LSTM, Temporal Convolutional Network (TCN), N-BEATS, Kolmogorov–Arnold Networks (KAN) and two quantum-inspired variants (QiLSTM, QiKAN). We describe dataset diagnostics, preprocessing and training recipes, and report aggregate point-forecast performance (MAE, RMSE) for every model. Our results show that methods encoding strong seasonal or low-dimensional functional structure perform particularly well on LMKS: Our quick-run results indicate that a quantum-inspired KAN variant (QiKAN) attains the lowest aggregate error in the evaluated configurations, while the simple Seasonal Naive baseline remains remarkably competitive. These findings highlight that, for highly periodic scientific monitoring series, parsimonious priors or decomposable function approximators can match or outperform more complex sequence models.
\end{abstract}
\textbf{Keywords.} neutron monitor, time-series forecasting, Kolmogorov--Arnold network, quantum-inspired models, LMKS

\section{Introduction}
Accurate and reliable time-series forecasting is a foundational capability for scientific monitoring systems. Ground-based neutron monitors such as the Lomnický Štít (LMKS) station produce continuous records of secondary cosmic-ray neutrons that are used in space-weather research, radiation-environment assessment, atmospheric studies and related operational tasks. These records are typically available at hourly (and sometimes sub-hourly) resolution and combine pronounced periodic behaviour (diurnal and seasonal cycles), long-range persistence, and intermittent transient excursions driven by solar and geomagnetic activity. Together, these characteristics make neutron monitor series an instructive and practically important testbed for multi-horizon forecasting methods.

From an operational perspective, reliable forecasts of neutron counts serve several concrete purposes: short- to medium-horizon prediction can support early warning systems for elevated radiation conditions, guide instrument scheduling and maintenance, and provide inputs to downstream decision-support algorithms. From a scientific perspective, forecasts and calibrated predictive intervals aid in separating predictable, quasi-periodic components from unusual transients that merit further investigation. In both use cases, models must balance sensitivity to transient events with robustness to regular seasonal structure, and they must supply well-calibrated uncertainty estimates to enable principled monitoring and alarm thresholds.

This work studies these challenges empirically by evaluating a diverse set of modelling approaches on the LMKS hourly record. We compare simple, interpretable baselines that explicitly encode seasonality with a range of learned sequence models that embody different inductive biases: memory-based recurrent models (LSTM), convolutional temporal models (TCN), interpretable basis-expansion networks (N-BEATS), Kolmogorov–Arnold inspired functional decompositions (KAN), and quantum-inspired variants that alter internal algebraic structure (QiLSTM, QiKAN). This diversity is intended to reveal which inductive biases most effectively capture the LMKS dynamics under limited tuning budgets and typical data-quality conditions.

\paragraph{Contributions.} The main contributions of this work are:
\begin{itemize}
  \item A reproducible multi-horizon forecasting benchmark on the Lomnick\'y \v{S}t\'it (LMKS) hourly neutron-monitor series, with preprocessing and training recipes.
  \item A comparative evaluation of seasonal baselines, LSTM, TCN, N-BEATS, KAN and quantum-inspired variants (QiLSTM, QiKAN) under a common training regime.
  \item Empirical evidence that parsimonious seasonality-aware and functional-decomposition models can match or outperform more complex sequence models on highly periodic monitoring series.
\end{itemize}

Note on terminology: by “quantum-inspired” we do not mean that these models were executed on quantum hardware. Instead the term denotes classical network variants that adopt algebraic motifs inspired by quantum-mechanical concepts (complex-valued states, phase-modulated recombination, approximate unitary constraints). These are classical, implementable architectures; any future hardware-based quantum implementation is left to future work.
\section{Related work}
Time-series forecasting has matured along several complementary directions, spanning classical statistical methods, tree-based learners \cite{breiman2001random,chen2016xgboost} and modern deep-learning architectures \cite{goodfellow2016deep}.  Recurrent networks such as LSTM and GRU remain fundamental for modelling temporal dependencies and gated memory \cite{hochreiter1997long}, while convolutional sequence models (notably the Temporal Convolutional Network) exploit dilated causal convolutions to achieve very large receptive fields with efficient parallelism \cite{bai2018empirical}.  Attention-based Transformers were introduced to capture flexible long-range interactions \cite{vaswani2017attention} and are widely used in NLP; for example BERT \cite{devlin2019bert}, and have been successfully adapted to forecasting problems where pairwise temporal relationships are important \cite{vaswani2017attention}.  Architectures such as N-BEATS \cite{oreshkin2019nbeats} adopt a different tack by using stacked fully-connected blocks with explicit backcast/forecast projections to provide an interpretable basis-expansion approach to multi-horizon forecasting.

A parallel strand of research studies architectures and representations motivated by classical function theory.  The Kolmogorov–Arnold representation theorem provides a theoretical basis for decomposing multivariate functions into sums of univariate components; Kolmogorov–Arnold Networks (KAN) operationalize this idea by combining learned linear projections with banks of univariate nonlinear subnetworks, a design that can be particularly parameter-efficient when the target mapping admits low-rank or decomposable structure \cite{kolmogorov1957,arnold1957,liu2024kan}.  Inspired by alternative algebraic structures, several recent works explore “quantum-inspired” modifications (e.g., QKAN \cite{ivashkov2024qkan}, QLSTM \cite{chen2020qlstm}) to classical architectures (for example complex-valued internal states, unitary-like transforms or phase-coupling mechanisms).  These variants aim to introduce different inductive biases that can improve expressivity in certain regimes, though they are often more sensitive to optimisation and hyperparameter choices.

On the uncertainty-quantification side, heteroscedastic Gaussian output parameterizations \cite{nix1994estimating} enable direct maximum-likelihood training of per-step variances \cite{nix1994estimating}, and deep ensembles provide a practical approach to capturing epistemic uncertainty \cite{lakshminarayanan2017simple}.  Conformal prediction complements model-native uncertainty by offering distribution-free recalibration procedures with finite-sample guarantees; recent work  \cite{gal2016dropout}, \cite{rasmussen2006gp}, \cite{kingma2014auto} has extended conformal ideas to multivariate and multi-horizon forecasting settings \cite{vovk2005algorithmic,angelopoulos2021gentle}.  Proper scoring rules such as CRPS, negative log-likelihood and classic point metrics (MAE, RMSE) remain standard for evaluating predictive quality and calibration \cite{gneiting2007strictly}.

Finally, for domain-specific application to particle accelerators and related scientific instrumentation, recent reviews synthesise methodological choices and practical considerations.  In particular, Li and Adelmann (2022) provide a focused review of time-series forecasting techniques and analyse their suitability for accelerator diagnostics and control tasks, highlighting practical trade-offs between interpretability, latency, and robustness in operational settings \cite{li2022review}.  Their survey situates the present study within the practical demands of scientific monitoring: short-latency forecasting, calibrated uncertainty for alarm thresholds, and robustness to instrumentation anomalies.

\section{Dataset: LMKS (Lomnický Štít Neutron Monitor)}
The Lomnický Štít neutron monitor (LMKS) provides one of the longest continuous ground-based records of cosmic-ray induced secondary neutrons. Its hourly series exhibits a combination of clear diurnal and seasonal cycles, long-range autocorrelations and occasional transient disturbances caused by solar and geomagnetic activity. These characteristics make LMKS an excellent testbed for benchmarking forecasting methods that must simultaneously capture regular periodicity and remain sensitive to rare events.

For this study we used the Lomnický Štít (LMKS) neutron-monitor release archived on Zenodo \cite{lmks2024zenodo} (Institute of Experimental Physics, Slovak Academy of Sciences; DOI: 10.5281/zenodo.10790916). Following the protocol implemented in our code, we first performed an audit of the raw data, removed non-numeric metadata fields, and applied forward/backward filling to handle missing hourly counts. All models were trained on standardized series segments to ensure comparability across architectures.

To transform the series into a supervised learning problem, we adopted a sliding-window approach: each training instance consists of a history of $T$ past values paired with the next $H$ target values. The quick-run configuration reported in this paper uses $T=20$ hours of history to forecast $H=10$ future hours, providing a balanced setting for assessing multi-horizon forecasting capability. This design yields a benchmark dataset that stresses both short-term predictive responsiveness and the ability to leverage longer-term seasonal structure.

\begin{figure}
    \centering
    \includegraphics[width=1\linewidth]{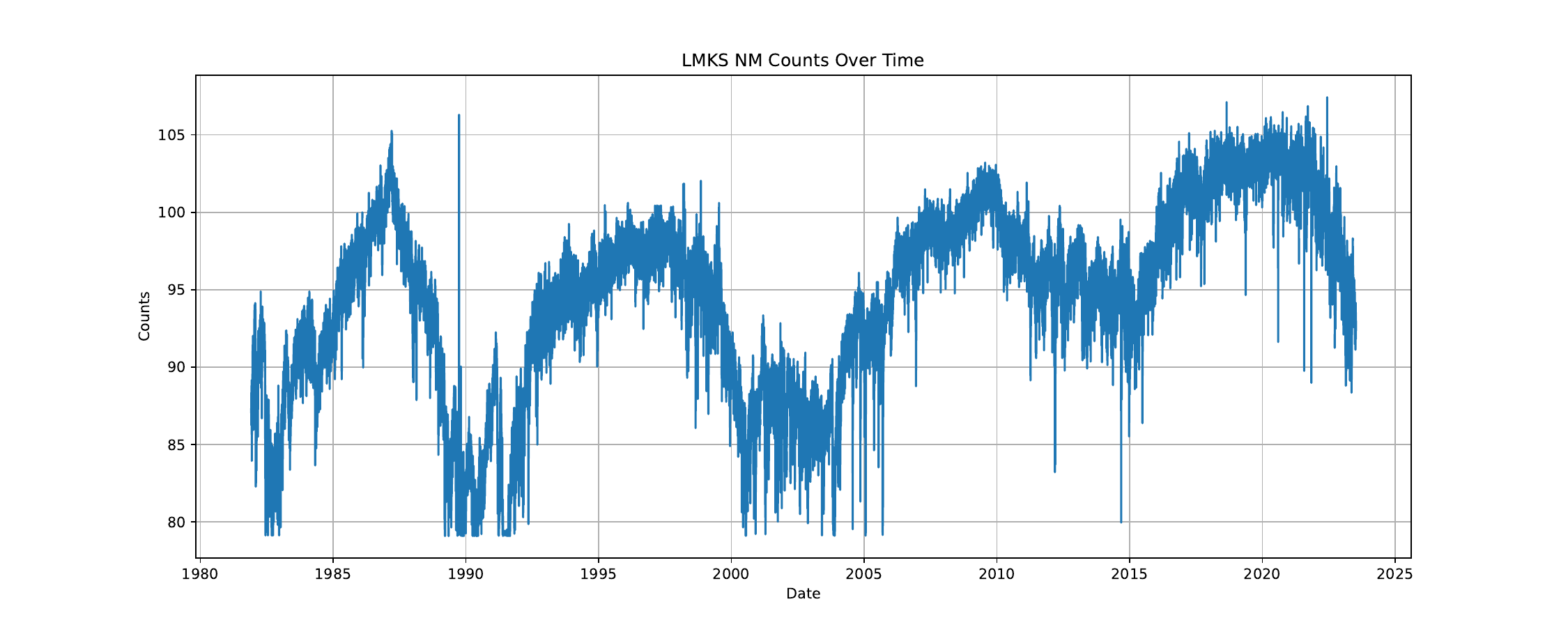}
    \caption{LMKS hourly neutron monitor time series. The series shows clear daily/seasonal patterns and transient variation that forecasting models must capture.}
    \label{fig:lmks_timeseries}
\end{figure}

We began with a feature audit and summary statistics. The LMKS series displays strong autocorrelations at short lags and pronounced seasonal structure at daily and longer scales. Missing observations are sporadic and were imputed with forward-fill followed by backward-fill; any leftover windows containing irrecoverable NaNs were dropped. For models requiring standardized inputs we computed per-channel training means and standard deviations and applied z-score normalization to train/validation/test splits. Figure~\ref{fig:lmks_timeseries} shows the hourly LMKS series used in our experiments.

\section{Models and implementation details}

\begin{figure}[H]
    \centering
    \includegraphics[width=0.6\linewidth, height = 7cm]{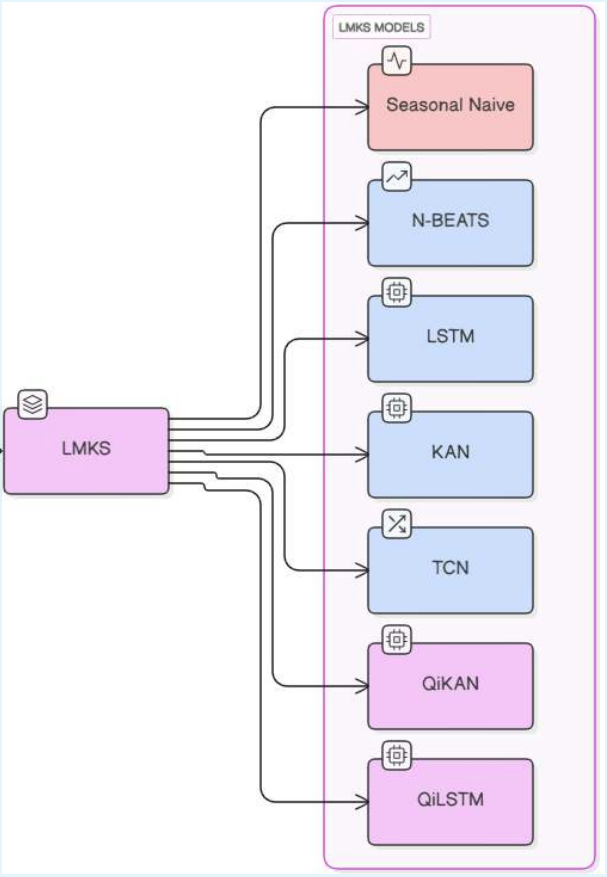}
    \caption{The data flow pipeline for the LMKS dataset. We analyse and forecast using different temporally designed models}
    \label{fig:dataflow}
\end{figure}

To ensure a reproducible and fair comparison, all models were implemented in PyTorch and trained under a common regime. Unless stated otherwise, we used the Adam optimiser with an initial learning rate of $10^{-3}$, batch size 64, gradient clipping at 1.0, and early stopping on validation mean squared error with a patience of eight epochs. Dropout between 0.1 and 0.3 was employed to regularise most models, and z-score normalisation based on training-set statistics was applied to all input sequences. Forecast quality was measured using both mean absolute error (MAE) and root mean squared error (RMSE). The overall pipeline is illustrated in Figure~\ref{fig:dataflow}.

\subsection{Seasonal Naive}
The Seasonal Naive predictor is an intentionally simple, fully deterministic baseline that directly copies observed seasonal structure into the forecast. Let $\{x_t\}$ denote the univariate time series sampled hourly and let $S$ denote the seasonal period (for the LMKS dataset, $S=24$ hours to reflect the daily cycle). For a forecast origin at time $t$ and horizon $h\in\{1,\dots,H\}$ the seasonal naive forecast is
\begin{equation}
\hat{x}_{t+h}^{\text{SN}} = x_{t+h-S}.
\label{eq:seasonal_naive}
\end{equation}
i.e. each predicted value is taken from the same hour in the previous daily cycle. Despite its simplicity, this rule is a strong benchmark for datasets with pronounced diurnal structure: it requires no parameter estimation, is entirely reproducible, and directly exposes the amount of residual structure that any learning model must explain beyond seasonality replication.

\subsection{Long-Short Term Memory (LSTM)}
The Long Short–Term Memory (LSTM) model is used in an encoder–decoder sequence-to-sequence formulation to transform a history window of length $T$ into a multi-step forecast of length $H$. Denote the input history by $\mathbf{x}_{t-T+1:t}=(x_{t-T+1},\dots,x_t)$. The encoder consumes this sequence and compresses temporal dependencies into a final latent state $\mathbf{z}_t$, while the decoder unfolds this representation into the horizon $\hat{\mathbf{x}}_{t+1:t+H}$. Internally the LSTM cell updates can be written (in standard real-valued form) as
\begin{equation}
\begin{aligned}
\mathbf{i}_u &= \sigma(W_i [\mathbf{h}_{u-1}, x_u] + b_i),\\
\mathbf{f}_u &= \sigma(W_f [\mathbf{h}_{u-1}, x_u] + b_f),\\
\mathbf{o}_u &= \sigma(W_o [\mathbf{h}_{u-1}, x_u] + b_o),\\
\tilde{\mathbf{c}}_u &= \tanh(W_c [\mathbf{h}_{u-1}, x_u] + b_c),\\
\mathbf{c}_u &= \mathbf{f}_u \odot \mathbf{c}_{u-1} + \mathbf{i}_u \odot \tilde{\mathbf{c}}_u,\\
\mathbf{h}_u &= \mathbf{o}_u \odot \tanh(\mathbf{c}_u).
\end{aligned}
\label{eq:lstm_cell}
\end{equation}
with $\sigma(\cdot)$ the logistic sigmoid and $\odot$ the elementwise product. In the encoder–decoder variant, the encoder produces a summary state that initializes the decoder; the decoder then produces outputs either autoregressively (feeding predicted values back as inputs) or with scheduled teacher forcing (mixing ground-truth and model predictions during training to stabilise learning). For deterministic forecasting the decoder's hidden states are mapped to scalar forecasts via a linear readout and trained under mean-squared error (MSE):
\begin{equation}
\mathcal{L}_{\text{MSE}} = \frac{1}{H}\sum_{h=1}^H \bigl(x_{t+h}-\hat{x}_{t+h}\bigr)^2.
\label{eq:mse}
\end{equation}
For probabilistic forecasting the model can be extended to predict heteroscedastic Gaussian parameters, producing a mean $\mu_{t+h}$ and a (positive) scale parameter $\sigma_{t+h}$; training then minimises the Gaussian negative log-likelihood,
\begin{equation}
\mathcal{L}_{\text{NLL}} = \frac{1}{2}\sum_{h=1}^H \left[ \log\!\big(\sigma_{t+h}^2\big) + \frac{\bigl(x_{t+h}-\mu_{t+h}\bigr)^2}{\sigma_{t+h}^2}\right] + \text{const}.
\label{eq:nll}
\end{equation}
where the variance is enforced positive via a smooth transform such as a softplus. Standard regularisation and optimisation practices — dropout between recurrent layers to reduce overfitting, and gradient clipping to avoid exploding gradients — support stable training and generalisation, while choices such as latent size or number of recurrent layers control the model’s capacity to capture longer-term temporal patterns.

\subsection{Temporal Convolutional Network (TCN)}
The Temporal Convolutional Network (TCN) offers a convolutional, non-recurrent approach to sequence modelling that retains causal structure and allows for very large receptive fields through dilated convolutions\cite{bai2018empirical}, a design popularized in WaveNet \cite{oord2016wavenet}. Let the convolutional kernel width be $k$ and suppose a stack of $L$ residual blocks is used with dilation factors $d_\ell$ (commonly $d_\ell = 2^{\ell-1}$). A single dilated causal convolution at layer $\ell$ computes
\begin{equation}
y^{(\ell)}_t = \sum_{m=0}^{k-1} w^{(\ell)}_m \, x_{t - d_\ell m}.
\label{eq:tcn_conv}
\end{equation}
guaranteeing that outputs at time $t$ depend only on past inputs. The effective receptive field of an exponentially dilated stack is
\begin{equation}
R = 1 + (k-1)\sum_{\ell=0}^{L-1} d_\ell.
\label{eq:tcn_receptive}
\end{equation}
which grows exponentially with depth and allows the model to cover the entire input window $T$ with relatively few layers. Each residual block typically contains two causal convolutions, nonlinearities (e.g. ReLU), dropout, layer normalisation and a residual connection that adds the block input to its output; this architecture preserves stable gradient flow and enables parallel prediction of the full horizon by mapping the final feature representation to the $H$-step forecast in one shot. The TCN is therefore well suited to efficient multi-step forecasting when the dependencies to be modelled are temporally local after dilation, and when autoregressive decoding is undesirable for latency reasons.

\subsection{N-BEATS}
N-BEATS is a purely feed-forward, block-stacked architecture that iteratively explains the input signal through repeated backcast subtraction while simultaneously contributing to the forecast. Given an input window $\mathbf{x}_{t-T+1:t}$, each block $b$ implements a function that outputs a backcast $\mathbf{b}^{(b)}$ (an approximation of the portion of the input it explains) and a forecast contribution $\mathbf{f}^{(b)}$:
\begin{equation}
\begin{aligned}
\big(\mathbf{b}^{(b)},\mathbf{f}^{(b)}\big) &= \mathcal{B}^{(b)}\big(\mathbf{r}^{(b)}\big),\\
\mathbf{r}^{(b+1)} &= \mathbf{r}^{(b)} - \mathbf{b}^{(b)}.
\end{aligned}
\label{eq:nbeats_block}
\end{equation}
with the residual $\mathbf{r}^{(1)}=\mathbf{x}_{t-T+1:t}$ and the final forecast produced by additive aggregation $\hat{\mathbf{x}}_{t+1:t+H} = \sum_{b} \mathbf{f}^{(b)}$. In the generic-block instantiation each $\mathcal{B}^{(b)}$ is a fully connected MLP that implicitly learns basis functions over the input window; the iterative backcast subtraction encourages blocks to specialise (for example into trend, seasonality or local corrections) even without an explicit decomposition. Training minimises the aggregated MSE over the horizon, and architectural choices such as block width, depth and number of blocks determine the richness of the learned basis expansions. The model’s design affords interpretability of block-level contributions while remaining flexible enough to approximate a wide class of temporal patterns.

\subsection{Kolmogorov–Arnold Networks (KAN)}
Kolmogorov–Arnold Networks (KAN) operationalise the classical Kolmogorov–Arnold representation theorem, which states that a continuous multivariate function can be expressed as a finite sum of univariate nonlinear functions applied to linear combinations of the inputs. Abstractly, for a mapping $F:\mathbb{R}^n\to\mathbb{R}$ there exist univariate functions $\phi_q$ and linear projection weights $a_{q}$ such that
\begin{equation}
F(\mathbf{x}) \approx \sum_{q=1}^{Q} \phi_q\big(\langle \mathbf{a}_q, \mathbf{x}\rangle\big).
\label{eq:kan_decomposition}
\end{equation}
A KAN model leverages this decomposition by first projecting the $T$-length history (viewed either as a $T$-dimensional vector or via a learned embedding) into a collection of scalar components $u_q = \langle \mathbf{a}_q, \mathbf{x}_{t-T+1:t}\rangle$, then passing each scalar $u_q$ through an independent univariate subnetwork $\phi_q$ and finally recombining the outputs linearly to form the multi-step forecast. In practice this induces a strong low-rank inductive bias: the function from high-dimensional history to forecast is modelled as a sum of simpler univariate nonlinearities, which yields parameter efficiency and can be particularly effective on quasi-periodic signals where dominant projections capture most of the explanatory variance. Regularisation, selection of $Q$, and the capacity of each $\phi_q$ control the trade-off between expressivity and overfitting.
\subsection{QiLSTM}
Quantum-inspired LSTM (QiLSTM) models extend the standard recurrent paradigm by permitting internal representations to carry phase information and by encouraging linear state transforms to behave like norm-preserving rotations. Concretely, the hidden state is treated as complex-valued, \(\mathbf{h}_u\in\mathbb{C}^d\), and recurrent updates are constructed so that the principal linear operators satisfy an approximate unitarity constraint \(W^\dagger W \approx I\). Gates and nonlinearities may be implemented either by operating separately on real and imaginary parts or by decomposing complex scalars into magnitude and phase and applying tailored operations to each component; final, real-valued predictions are obtained by mapping complex outputs to the real domain via the real part, the magnitude, or a learned linear readout that consumes concatenated real and imaginary channels. Practically, a structured front end that produces compact, phase-aware embeddings of the input history is often used: local scalar encodings are combined through a chain of small linear cores whose inputs are formed by the outer-product between the current bond state and the local encoding (flattened and passed through the core), and the resulting final bond state is pooled and projected to form the LSTM input. Because QiLSTM aims to represent sustained oscillations and phase-locked phenomena, training typically includes stability priors such as a unitarity regularizer
\begin{equation}
\mathcal{R}_{\mathrm{unit}}(W)=\bigl\lVert W^\dagger W - I\bigr\rVert_F^2.
\label{eq:unitarity_reg}
\end{equation}
and optional amplitude or phase-drift penalties; alternatively, one may parameterize certain linear blocks to be exactly unitary (e.g., via exponentiation of skew-Hermitian generators or structured products of Givens rotations) when exact norm preservation is required. Together, the complex-valued state, phase-aware nonlinearities, and norm-preserving linear components give QiLSTM a compact mechanism for encoding rotations, phase delays, and interference-like combination rules that are especially effective when the target dynamics contain prominent frequency and phase structure.

\subsection{QiKAN}
The quantum-inspired Kolmogorov–Arnold Network (QiKAN) adapts the KAN decomposition to support phase and interference by combining many localized univariate transforms with a small number of learned aggregators. Given an input vector \(x=(x_1,\dots,x_n)\), QiKAN applies per-coordinate univariate subnetworks \(\phi_j:\mathbb{R}\to\mathbb{C}^{m}\) (or \(\mathbb{R}^m\) when a real formulation is preferred) and forms \(K\) aggregator channels whose inputs are linear combinations of these univariate encodings:
\begin{equation}
u_k(x)=\sum_{j=1}^n a_{kj}\,\phi_j(x_j),\quad k=1,\dots,K.
\label{eq:qikan_aggregator}
\end{equation}
where the coefficients \(a_{kj}\) are learned and may be complex to permit constructive and destructive interference across coordinates. Each aggregator output \(u_k\) is then mapped through a small nonlinear map \(h_k\) to produce a contribution, and the final prediction is the sum of these contributions,
\begin{equation}
\hat{f}(x)=\sum_{k=1}^K h_k\big(u_k(x)\big).
\label{eq:qikan_prediction}
\end{equation}
This architecture realizes a low-rank, compositional recombination of localized features: when cross-coordinate interactions are structured or effectively low-dimensional, a modest number of aggregators suffices to capture global nonlinear effects while remaining statistically efficient. In the complex-valued variant, relative phases in \(\phi_j(x_j)\) and in the coefficients \(a_{kj}\) enable interference patterns that can succinctly represent oscillatory coupling across inputs; to preserve numerical stability one therefore typically couples the QiKAN with the same regularization motifs used in QiLSTM (unitarity or approximate norm-preservation on linear recombination layers, amplitude/phase penalties, and spectral normalization). Training objectives combine the usual predictive loss (e.g., MSE or MAE) with these structural regularizers, and practical implementations often fit any dimensionality-reducing transforms on training data only, checkpoint on validation performance, and tune the strength of unitary/phase penalties so as to stabilize dynamics without removing useful amplitude modulation.


\subsection{Calibration and evaluation metrics}
Model performance was primarily assessed using point-forecast errors, measured by Mean Absolute Error (MAE) and Root Mean Squared Error (RMSE). For a dataset of $N$ forecast instances, each with a horizon of length $H$, and predictions $\hat{y}_{i,t}$ against ground truth $y_{i,t}$, the metrics are defined as
\begin{equation}
\text{MAE} \;=\; \frac{1}{N H} \sum_{i=1}^{N} \sum_{t=1}^{H} \bigl| y_{i,t} - \hat{y}_{i,t} \bigr| .
\label{eq:mae}
\end{equation}

\begin{equation}
\text{RMSE} \;=\; \sqrt{ \frac{1}{N H} \sum_{i=1}^{N} \sum_{t=1}^{H} \bigl( y_{i,t} - \hat{y}_{i,t} \bigr)^2 } \, .
\label{eq:rmse}
\end{equation}

These definitions match the implementation used in our codebase, where errors are first computed element-wise across all horizon steps and then averaged across the full evaluation set.

\section{Results}
Table \ref{tab:lmks_results} summarises aggregate forecasting performance (MAE and RMSE) for all evaluated models on the LMKS dataset. The numbers are aggregated across the chosen horizon and were produced under the execution configuration described above.

\begin{table}[htbp]
\centering
\caption{LMKS forecasting results: aggregate MAE and RMSE across the forecast horizon (quick-run snapshot).}
\label{tab:lmks_results}
\begin{tabular}{lrr}
\toprule
Model & MAE & RMSE \\
\midrule
Seasonal Naive & 0.5502 & 0.8503 \\
LSTM           & 0.5763 & 0.8111 \\
TCN            & 0.8194 & 1.1288 \\
N-BEATS        & 0.5847 & 0.8095 \\
KAN            & 0.5831 & 0.8125 \\
QiLSTM         & 0.7739 & 1.0020 \\
QiKAN          & 0.5276 & 0.7896 \\
\bottomrule
\end{tabular}
\end{table}

\begin{figure}[H]
    \centering
    \includegraphics[width=1.0\linewidth]{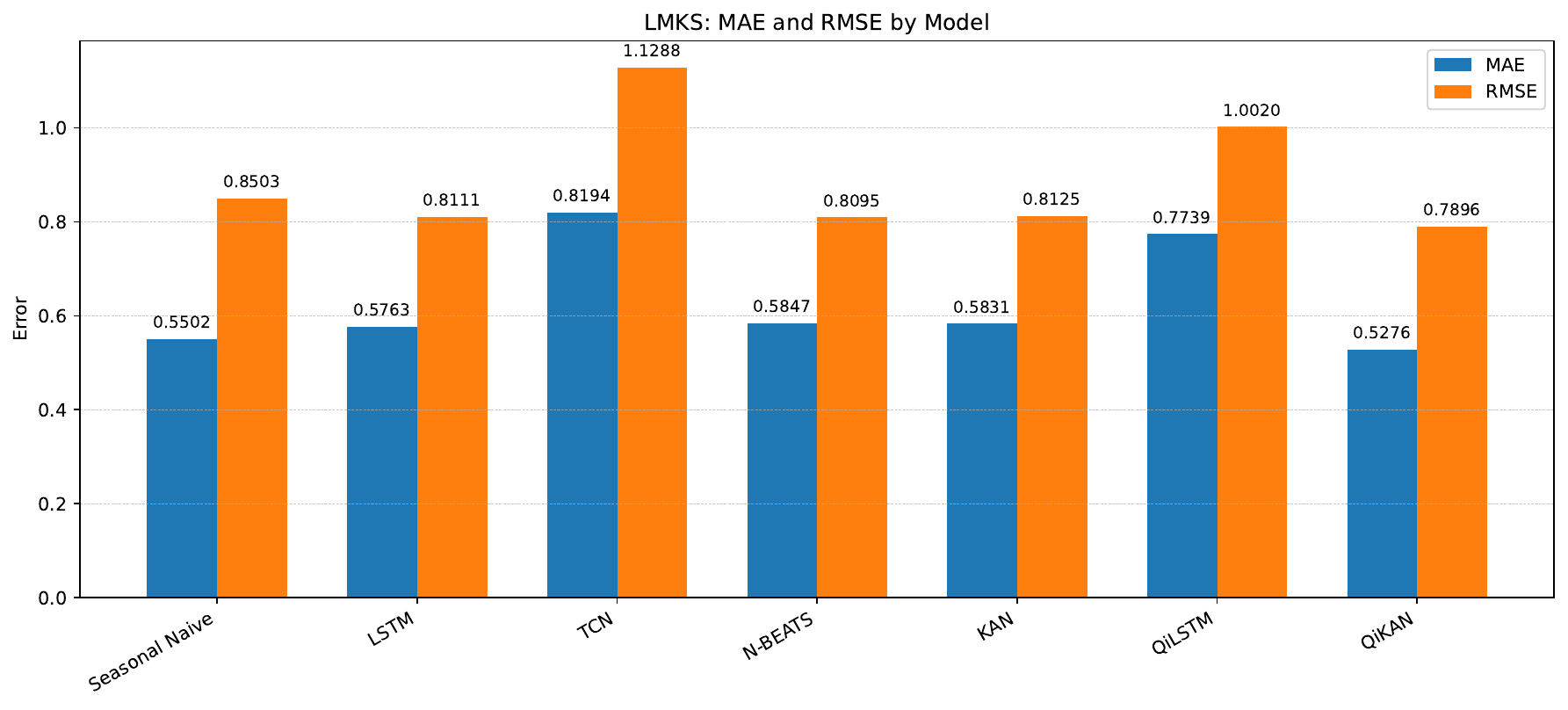}
    \caption{Results for the LMKS dataset. In the quick-run evaluation reported here, the quantum-inspired KAN variant (QiKAN) achieved the lowest aggregate point forecast error among the configurations tested.}
    \label{fig:lmks_metrics}
\end{figure}

\subsection*{Analysis}
The quantum-inspired KAN variant (QiKAN) achieves the lowest aggregate errors in this quick-run snapshot. This outcome indicates that strong periodic priors and low-dimensional functional decompositions are particularly effective on the LMKS series, which exhibits pronounced periodicity and slowly varying behavior. Classical sequence models (LSTM) and functional deep models (N-BEATS, KAN) produce similar error magnitudes, while the TCN underperformed in our default quick-run hyperparameter choices — likely because its receptive-field design and convolutional inductive bias require careful kernel/dilation tuning to match the LMKS seasonality scale. Quantum-inspired LSTM (QiLSTM) shows higher error in the quick-run setting, suggesting optimisation sensitivity that merits more exhaustive tuning. Summary metrics are plotted in Figure~\ref{fig:lmks_metrics}.

\paragraph{Limitations.}
This study used a focused, quick-run experimental protocol with a modest hyperparameter search budget; as a consequence, some models (notably the Qi-variants and TCN) exhibited sensitivity to initialization and regularisation choices. Results reported here should therefore be interpreted as a reproducible snapshot under the stated defaults rather than exhaustive performance bounds. We evaluated a single dataset snapshot (LMKS hourly record from 1981-12-01 through 2023-07-10); generalisation to other neutron-monitor series or different sampling resolutions may require further tuning and validation. Future work will expand hyperparameter sweeps, report full seed-averaged statistics, and provide the full code and checkpoint artifacts for independent verification.

\section{Discussion and Conclusion}
Several practical lessons emerge from the LMKS experiments. First, when a time series shows a strong, regular seasonal pattern, simple seasonality-aware baselines—seasonal naïve forecasts, classical decomposition (trend + seasonal + residual), or lightweight parametric seasonality models—often perform very competitively. These methods exploit dominant periodicity directly, remain stable over medium-length horizons, and require minimal hyperparameter tuning or large training sets; for monitoring tasks with clear periodicity they are reliable, low-cost baselines.

Second, modern deep sequence models provide considerable representational flexibility but can be fragile with respect to architecture and receptive-field choices. Temporal Convolutional Networks (TCNs), Transformers, and related architectures can outperform classical methods, but only when receptive fields, positional encodings, dilation patterns, attention windows, and other design choices are well matched to the data’s periodicity and the forecast horizon. In our LMKS experiments—where seasonality is strong and horizons are moderate—TCNs and Transformers required careful configuration to consistently beat simpler baselines, underscoring the need for targeted architecture search and thorough validation.

Third, quantum-inspired and quantum-hybrid architectures are an intriguing exploratory direction. Our quick-run results show promise but indicate these models are not yet plug-and-play: they demand careful optimisation, appropriate inductive biases, and architecture-specific regularisation to reliably surpass classical baselines. Practical considerations—parameter initialisation, constrained parameter counts, robustness to noise, and training schedules—strongly influence whether a quantum-inspired variant can deliver its theoretical advantages in finite-data, noisy settings.

These conclusions come from a focused evaluation on the LMKS neutron monitor dataset. By comparing seasonal baselines, recurrent and convolutional sequence models, interpretable basis-expansion networks (functional decomposition), and quantum-inspired variants, we present a practical landscape of model performance for scientific monitoring. The clearest takeaway is the enduring strength of seasonality-aware baselines and functional decomposition: they yield dependable results with modest complexity, while more complex models require careful tuning to justify their use.

Future work will implement and evaluate Variational Quantum Circuit (VQC)–based hybrid architectures—specifically a quantum-augmented LSTM and a quantum KAN—by replacing or augmenting key linear transforms with compact VQC modules and by exploring encodings (angle, amplitude, IQP-style) suited to periodic signals. Experiments will proceed from high-fidelity simulators to hardware runs, employ staged training and parameter-efficient ansätze, and use architecture-specific regularisation; benchmarks, ablations, and reproducible code, seeds, and configurations will clarify whether VQC hybrids offer consistent expressivity or robustness benefits for scientific monitoring.
%
%

\end{document}